\documentclass[sigconf]{acmart}

\usepackage{balance}
\AtBeginDocument{%
  \providecommand\BibTeX{{%
    \normalfont B\kern-0.5em{\scshape i\kern-0.25em b}\kern-0.8em\TeX}}}

\copyrightyear{2021} 
\acmYear{2021} 
\setcopyright{acmcopyright}
\acmConference[MuSe '21] {Proceedings of the 2nd Multimodal Sentiment Analysis Challenge}{October 24, 2021}{Virtual Event, China.}
\acmBooktitle{Proceedings of the 2nd Multimodal Sentiment Analysis Challenge (MuSe '21), October 24, 2021, Virtual Event, China}
\acmPrice{15.00}
\acmISBN{978-1-4503-8678-4/21/10} 
\acmDOI{10.1145/3475957.3484457}

\usepackage{array}
\newcommand{\PreserveBackslash}[1]{\let\temp=\\#1\let\\=\temp}
\newcolumntype{C}[1]{>{\PreserveBackslash\centering}p{#1}}
\newcolumntype{R}[1]{>{\PreserveBackslash\raggedleft}p{#1}}
\newcolumntype{L}[1]{>{\PreserveBackslash\raggedright}p{#1}}

\begin{document}
\fancyhead{}

\title{Hybrid Mutimodal Fusion for Dimensional Emotion Recognition }




\author{Ziyu Ma}
\authornote{Both authors contributed equally to this research.}
\affiliation{%
 \institution{College of Electrical and Information Engineering, \\Hunan University}
 \institution{Key Laboratory of Visual Perception and Artificial Intelligence of Hunan Province}
 \city{Changsha}
 \country{China}}
\email{maziyu@hnu.edu.cn}

\author{Fuyan Ma}
\authornotemark[1]
\affiliation{%
 \institution{College of Electrical and Information Engineering, \\Hunan University}
 \institution{Key Laboratory of Visual Perception and Artificial Intelligence of Hunan Province}
 \city{Changsha}
 \country{China}}
\email{mafuyan@hnu.edu.cn}

\author{Bin Sun}
\affiliation{%
 \institution{College of Electrical and Information Engineering, \\Hunan University}
 \institution{Key Laboratory of Visual Perception and Artificial Intelligence of Hunan Province}
 \city{Changsha}
 \country{China}}
\email{sunbin611@hnu.edu.cn}

\author{Shutao Li}
\affiliation{%
 \institution{College of Electrical and Information Engineering, \\Hunan University}
 \institution{Key Laboratory of Visual Perception and Artificial Intelligence of Hunan Province}
 \city{Changsha}
 \country{China}}
\email{shutao\_li@hnu.edu.cn}


\begin{abstract}
In this paper, we extensively present our solutions for the MuSe-Stress sub-challenge and the MuSe-Physio sub-challenge of Multimodal Sentiment Challenge (MuSe) 2021.
The goal of MuSe-Stress sub-challenge is to predict the level of emotional arousal and valence in a time-continuous manner from audio-visual recordings and the goal of MuSe-Physio sub-challenge is to predict the level of psycho-physiological arousal from a) human annotations fused with b) galvanic skin response (also known as Electrodermal Activity (EDA)) signals from the stressed people.
The Ulm-TSST dataset which is a novel subset of the audio-visual textual Ulm-Trier Social Stress dataset that features German speakers in a Trier Social Stress Test (TSST) induced stress situation is used in both sub-challenges.
For the MuSe-Stress sub-challenge, we highlight our solutions in three aspects:
1) the audio-visual features and the bio-signal features are used for emotional state recognition.
2) the Long Short-Term Memory (LSTM) with the self-attention mechanism is utilized to capture complex temporal dependencies within the feature sequences.
3) the late fusion strategy is adopted to further boost the model's recognition performance by exploiting complementary information scattered across multimodal sequences.
Our proposed model achieves CCC of 0.6159 and 0.4609 for valence and arousal respectively on the test set, which both rank in the top 3. For the MuSe-Physio sub-challenge, we first extract the audio-visual features and the bio-signal features from multiple modalities. Then, the LSTM module with the self-attention mechanism, and the Gated Convolutional Neural Networks (GCNN) as well as the LSTM network are utilized for modeling the complex temporal dependencies in the sequence. Finally, the late fusion strategy is used. Our proposed method also achieves CCC of 0.5412 on the test set, which ranks in the top 3.

\end{abstract}




\maketitle

\section{Introduction}
The emotional states play a crucial role in human work and daily life through affecting psychological and physiological status. It is noted that many people in modern societies are under a high level of stress \cite{1}. What's worse, the long-term accumulation of negative emotions may result in depression, which has an extremely severe impact on mental health. Thus, accurately recognizing human emotions is a fundamental but profound research direction. Emotion recognition systems have various applications, such as driver fatigue monitoring \cite{2}, health care especially mental health monitoring \cite{3}, building a secure and safe society \cite{4} and human-robot interaction.\par

Although the discrete categorical theory of emotion is a popular computing model for facial expression recognition \cite{ma2021robust}, the continuous dimensional theory of emotion is the most commonly used for emotion prediction in time-continuous sequences.
According to \cite{fontaine2007world}, many researchers have focused on dimensional models, such as the valence-arousal model\cite{yik1999structure}.
For example, Russel et al. \cite{5} proposed a computing model of emotions characterized by two essential dimensions (i.e., valence and arousal). Marsella et al. \cite{6} developed a model that treated a specific emotional state as a data point in the continuous space, which was described mathematically by 3 dimensions corresponding to arousal (a measure of affective activation), valence (a measure of pleasure), and dominance (a measure of control). Therefore, the continuous dimensional theory of emotion is suitable to compute the sentiments scattered across multi-modal sequences, because it can describe more subtle, continuous and complicated emotional states. \par

As we know, it is significantly subjective to annotate the emotions of human beings. However, multiple professional annotators must continuously annotate multi-modal sequences, which is time-consuming and laborious. Therefore, how to produce an agreed-upon representation (gold standard) from multiple labelers still remains an open research question, with few methods available. For example, given the level of disagreement, Michael et al. \cite{7} proposed to use the Evaluator Weighted Estimator to weight different annotators based on the likelihood of agreement. It is limited for the research on the fusion of perceived emotional signals with extra physiological signals and there also has been minimal research on the combination between physiological and perceived arousal gold standard. Therefore, in the 2021 edition of the Multi-modal Sentiment in-the-wild (MuSe) challenge \cite{muse} , apart from predicting the level of emotional arousal and valence in a time-continuous manner from people in stressed dispositions for the MuSe-Stress sub-challenge, the signal of arousal is fused with the Electrodermal Activity (EDA) and then used as a prediction target for the MuSe-Physio sub-challenge.
Participants are provided with the multimodal Ulm-TSST data-base, where subjects were recorded under a highly stress-induced free speech task.\par

In the MuSe-Stress sub-challenge, eGeMAPS is the best audio feature for the prediction of valence and arousal, achieving 0.5018 CCC and 0.4416 CCC on the test set respectively which reported in the baseline paper. VGGface is the best vision feature for the prediction of valence and arousal, achieving 0.4529 CCC and 0.1579 CCC on the test set respectively which reported in the baseline paper. Therefore, we choose eGeMAPS and VGGface feature sets for the MuSe-Stress sub-challenge in this paper.
Apart from the audio features (eGeMAPs), the visual features (VGGface), the bio-signal features (Electrocardiogram (ECG), Respiration (RESP), and heart rate (BPM)) are also selected. We apply the Long Short-Term Memory (LSTM) with the self-attention module for modeling the complex temporal dependencies in the sequence from the audio-visual modalities and the LSTM network is used to capture the complex and continuous dependencies within the sequences from bio-signal modality. Finally, we concatenate all multi-modal features and subsequently send them into the regression model.\par

In the MuSe-Physio sub-challenge, the late fusion of the best audio (VGGish) and video (VGGface) predictions yield the best results, namely 0.4913 CCC on development data and 0.4908 CCC on test data according to the baseline paper. Therefore, we choose VGGish and VGGface feature sets for the MuSe-Physio sub-challenge in this paper.
In addition, the bio-signal features (ECG, RESP, and BPM) are also the input of the model. Then, we utilize the LSTM with the self-attention mechanism module and the Gated Convolutional Neural Networks (GCNN) as well as the LSTM network for modeling the complex temporal dependencies in the sequence from the audio-visual modality respectively. The bio-signal features are concatenated and sent to the LSTM network. Finally, we concatenate all types of multi-modal features before sending them into the regression model.

Our main contributions to the MuSe 2021 challenge in this paper can be summarized as follows:
\begin{itemize}
\item {\verb||}
  We investigate the impact of various features including audio features, visual features and bio-signal features in both sub-challenges. We also demonstrate the effectiveness of the bio-signal features, which has not been well-used in corresponding sub-challenges of previous MuSe.
\item{\verb||}
We also propose to augment the LSTM network with the self-attention mechanism for continuous emotion prediction. The results demonstrate the combination of these two modules can enhance the ability of  modeling long-term temporal dependencies within the sequence. Moreover, with the help of the late fusion technique, our method achieves promising results in the MuSe-Stress sub-challenge.
\item{\verb||}
The LSTM network enhanced with the self-attention mechanism and Gated CNN-LSTM model is proposed for further extracting the information from audio-visual modalities. The late fusion technique is also used as the MuSe-Stress sub-challenge and our model achieves competitive results in the MuSe-Physio sub-challenge.
\end{itemize}

The rest of this paper is structured as follows: Section 2 introduces the related work. The multimodal features are presented in Section 3. The methods are described in Section 4. The experimental setting and the results are presented in Section 5. Finally, we make the conclusive remarks and indicates the possible future directions in Section.

\section{Related Work}
{\bfseries Multi-modal Features}: Various multimodal features have been well investigated in the past series of AVECs, which are the predecessor of MuSe.
Before the deep learning era, competitors tended to use carefully-handcrafted features for regression. For example, S{\'a}nchez-Lozano et al. \cite{8} proposed to extract the Gabor feature and the Local Binary Patterns (LBP) for the individual visual modality and extract low-level features for the single audio modality. With the deep learning technique showing superior performance over carefully-handcrafted features, participants have been more likely to learn deep representations for both modalities by different variants of very deep networks. The methods \cite{9} \cite{10} in AVEC 2017 demonstrated that utilizing the convolutional neural networks for the visual representations obtained comparable and even better results than carefully-engineered features. Furthermore, the results of \cite{11} in AVEC 2018 demonstrated that the learned audio representations from the pretrained VGGish model could outperform expert-knowledge based acoustic features. Chen et al. \cite{12} proposed to use 2D+1D convolutional neural networks in the very recent AVEC 2019, which also proved the superior performance of learned audio-visual features. Apart from audio-visual modalities, the linguistic features play an important part in multimodal emotion analysis \cite{13}. In AVEC 2017, the textual modality was first introduced to solve the recognition problem. At the very beginning, participants used the classic bag-of-words features \cite{10}. Subsequently, the word vectors (such as Word2Vec \cite{14} and GloVe \cite{15}) were widely applied because of their efficiency and effectiveness, which were pre-trained on large-scale text datasets.

{\bfseries Model Architecture}: Support vector regression (SVR) is the most commonly used as the baseline methods in previous AVECs. However, because of its non-temporal disadvantage, SVR cannot use the temporal and contextual information well to promote the performance of emotion recognition and sentiment analysis. With the increasing popularity of deep neural networks, the Recurrent Neural Network (RNN) has outperformed better in sequence modeling over other methods. Therefore, the 1st place winners of recent AVECs all adopted the LSTM network without exception, for continuous sentiment analysis. Previous methods \cite{16}, \cite{9} both utilized the LSTM and SVR to perform emotion regression. The experimental results demonstrated that the LSTM outperformed SVR dramatically. Compared with RNN based methods, a fully convolutional network was designed by Du et al. \cite{17}. It performed emotion recognition in a coarse-to-fine strategy by aggregating multi-level features with various scales. Huang et al. \cite{18} utilized various types of temporal architectures to capture temporal contextual relationships within sequences. Their studies also revealed that the combination of various models resulted in better results.

{\bfseries Multi-modal Fusion}: 
It is indispensable for researchers to utilize multi-modal fusion to boost the performance of emotion prediction systems. Huang et al. \cite{10} adopted the late fusion technique to aggregate the predictions based on different types of features in AVEC 2017, while the early fusion strategy was used by Chen et al. \cite{9} to capture the inter-modal dynamics. In AVEC 2018, the detail comparison results of these methods were described by Huang et al. \cite{19}. The comparison demonstrated that methods with the late fusion predicted arousal and valence better than these with the early fusion. In AVEC 2019, Chen et al. \cite{20} proposed to utilize both the early and the late fusion techniques in one model. They first trained the deep Bidirectional Long Short-Term Memory networks (Bi-LSTM) for unimodal features. Afterwards, Bi-LSTMs were applied for fusing corresponding bi-modal features in an early stage. Finally, the predictions of various models were fused late by a second level Bi-LSTM. Moreover, Huang et al. \cite{21} proposed to fuse multimodal information by the cross-modal attention module. Their results showed the proposed cross-modal attention could outperform better over both the early and the late fusion techniques.

\section{Multimodal Features}

\subsection{Visual Features}
{\bfseries VGGface Feature}: 
The visual geometry group of Oxford introduced the deep CNN referred to as VGG16 \cite{25}.
As a variant of VGG16, the VGGface architecture was initially intended for supervised facial recognition purposes \cite{25}, which was trained on 2.6 million faces.
To get rid of irrelevant background information, we first apply MTCNN to detect and align faces in videos.
VGGface \cite{28} is then used to extract general facial features, which can present high-level and discriminative features compared with other facial recognition models.

\subsection{Acoustic Features}
{\bfseries eGeMAPS Feature}: 
We use the extended Geneva Minimalistic Acoustic Parameter Set (eGeMAPS) feature provided by the organizers of MuSe 2021, which contains 23 acoustic low-level descriptors (LLDs \cite{22}. The freely and publicly available openSMILE toolkit \cite{23} can be used to extract the eGeMAPS feature. Several statistical functions of the openSMILE toolkit can be directly applied to extract segment-level features with an 88-dimensional vector.

{\bfseries VGGish Feature}: VGGish \cite{24} was pre-trained on AudioSet \cite{26}.
Although VGGish was originally proposed for audio classification, previous studies, such as \cite{27}, demonstrated that the learned deep feature representation from trained VGGish outperformed carefully-handcrafted acoustic features.
Therefore, the pre-trained VGGish model is used to extract acoustic features in our work.
We first divided the recordings into multiple 0.975 s frames with a hop size of 0.25s to match with the ground-truth labels. Afterwards, we extract the log spectrograms from these frames. Subsequently, they are sent into the VGGish model. The 128-dimensional embeddings from the output of fc2 layer are extracted as the final VGGish feature.

\subsection{Biological Features}
{\bfseries ECG, RESP and  BPM Feature}: 
Three biological signals (i.e., Electrocardiogram (ECG), Respiration (RESP), and heart rate (BPM)) were captured at a sampling rate of 1 kHz. We use the official biological features at a sampling rate of 2Hz as the inputs for our model. It is noted that the bio-signals are newly featured for MuSe 2021, which has not been well investigated for continuous dimensional emotion recognition.

\section{Methods}
\subsection{MuSe-Stress sub-challenge}
In this sub-challenge, the LSTM networks with the self-attention mechanism is utilized to capture complex temporal dependencies within the feature sequences. Different from \cite{sun} ,  the bio-signal features (Electrocardiogram (ECG), Respiration (RESP), and heart rate (BPM)) are also selected.
Figure 1 demonstrates the architecture of our model, which contains three modules: the LSTM networks with the self-attention module for the acoustic and visual features, the LSTM module for the bio-signal features and the late fusion module. First, we feed the eGeMAPS feature and the VGGface feature to the LSTM network with self-attention module respectively and obtain the predictions from both audio and visual modalities. Then, the bio-signal features are concatenated and sent to the LSTM module in order to get the predictions of bio-signal modality. Finally, the predictions from above modalities are concatenated and sent to the late fusion module for regression.

\begin{figure}[t]
  \centering
  \includegraphics[width=85mm]{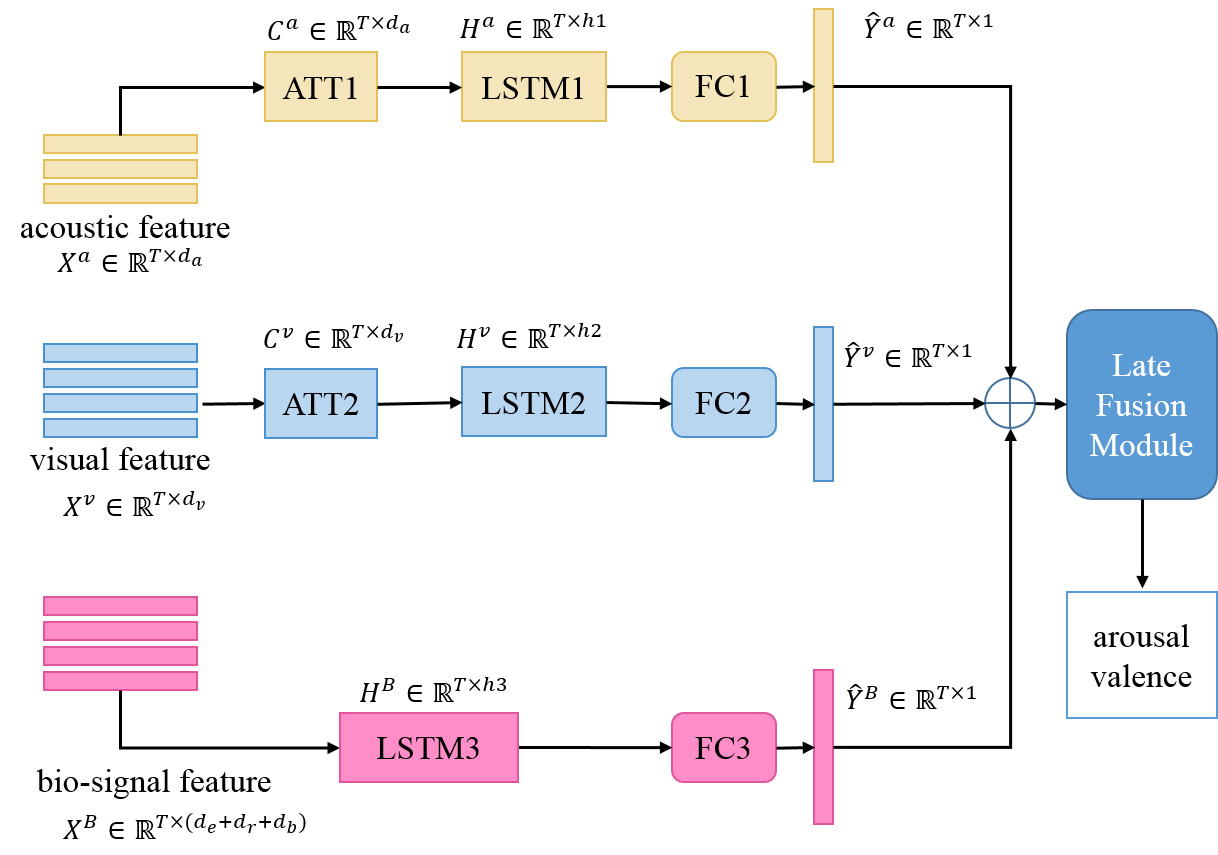}
  \caption{Overview of the architecture used in the MuSe-Stress sub-challenge. It contains three modules: the LSTM networks with the self-attention module for the acoustic and visual features, one LSTM module for the bio-signal features and the late fusion module.}
  \Description{}
\end{figure}

{\bfseries Self-attention Mechanism}: The self-attention mechanism is utilized to transform the input sequences into high-level representations, which further captures the relationships across the sequences.
Given an input acoustic sequence $\boldsymbol{X}^{a}=\left\{\boldsymbol{X}_{i}^{a} \in \mathbb{R}^{d_{a}}| i=1,...,|T|\right\}$, the self-attention output $\boldsymbol{C}^{a}=\left\{\boldsymbol{C}_{i}^{a} \in \mathbb{R}^{d_{a}}| i=1,...,|T|\right\}$ is given by:
{
\begin{equation}\label{1}
\boldsymbol{C}^{a}=\operatorname{softmax}\left(\frac{\mathbf{{Q}^{a} {K}^{a}}^{\top}}{\sqrt{d_{k}}}\right) \mathbf{{V}^{a}}
\end{equation}
where $d_{a}$ is the dimension of the acoustic sequence and $|T|$ is the max time step. $\mathbf{{Q}^{a}}$, $\mathbf{{K}^{a}}$ and $\mathbf{{V}^{a}}$ represent the queries, keys and values matrices respectively which mapped by the input acoustic sequence $\boldsymbol{X}^{a}$. $d_{k} = d_{a} / h$ is the scale factor and $h$ represents the number of heads. For simplicity, we override the whole operation as a mapping $\operatorname{ATT}$:
\begin{equation}\label{2}
\boldsymbol{C}^{a}= \operatorname{ATT_{1}}(\boldsymbol{X}^{a})
\end{equation}
}
In our model, the visual feature is also sent to the self-attention module to capture the contextual information. Similarly, given an input visual sequence $\boldsymbol{X}^{v}=\left\{\boldsymbol{X}_{i}^{v} \in \mathbb{R}^{d_{v}}| i=1,...,|T|\right\}$, the self-attention output $\boldsymbol{C}^{v}=\left\{\boldsymbol{C}_{i}^{v} \in \mathbb{R}^{d_{v}}|i=1,...,|T|\right\}$ is calculated by:
{
\begin{equation}\label{3}
\boldsymbol{C}^{v}= \operatorname{ATT_{2}}(\boldsymbol{X}^{v})
\end{equation}
where $d_{v}$ is the dimension of the visual sequence.
}

{\bfseries LSTM}: In our model, the outputs of the self-attention modules are sent to the LSTM to capture the complex temporal dependencies within the sequence. Given the input embedding sequence $\boldsymbol{C}^{a}$, the hidden semantic feature is repeatedly computed at the $i$-th
step by:
{
\begin{equation}\label{4}
\overrightarrow{\boldsymbol{H}}_{i}^{a}=f_{L S T M}( \boldsymbol{C}_{i}^{a}, \overrightarrow{\boldsymbol{H}}_{i-1}^{a})
\end{equation}
where $f_{L S T M}$ represents the LSTM function. For simplicity, we override the whole operation as a mapping $\operatorname{L S T M_{1}}$ :

\begin{equation}\label{5}
(\boldsymbol{H}_{1}^{a}, \ldots, \boldsymbol{H}_{|T|}^{a}) = \operatorname{L S T M}_{1}(\boldsymbol{C}_{1}^{a}, \ldots, \boldsymbol{C}_{|T|}^{a})
\end{equation}
where $\boldsymbol{H}_{i}^{a} \in \mathbb{R}^{h_{1}}$ and $h_{1}$ represents the dimension of $\operatorname{L S T M_{1}}$. Similarly, given the input embedding sequence $\boldsymbol{C}^{v}$, the LSTM output is calculated by:
\begin{equation}\label{6}
(\boldsymbol{H}_{1}^{v}, \ldots, \boldsymbol{H}_{|T|}^{v}) = \operatorname{L S T M}_{2}(\boldsymbol{C}_{1}^{v}, \ldots, \boldsymbol{C}_{|T|}^{v})
\end{equation}
where $\boldsymbol{H}_{i}^{v} \in \mathbb{R}^{h_{2}}$ and $h_{2}$ represents the dimensionality of $\operatorname{L S T M_{2}}$. Finally, $\boldsymbol{H}^{a}$ and $\boldsymbol{H}^{v}$ are sent to the fully connected layers to obtain the emotion predictions respectively. For the audio modality, the emotion prediction $\hat{Y}^{a}=\left\{\hat{Y}_{1}^{a}, \ldots, \hat{Y}_{|T|}^{a}\right\}$ is given by 
\begin{equation}\label{7}
\hat{Y}^{a}=\boldsymbol{H}^{a} \boldsymbol{W}^{a}+ \boldsymbol{b}^{a}
\end{equation}
where $\boldsymbol{W}^{a} \in \mathbb{R}^{h1 \times 1}$, and $\boldsymbol{b}^{a}$ represents the bias. For the video modality, the emotion prediction $\hat{Y}^{v}=\left\{\hat{Y}_{1}^{v}, \ldots, \hat{Y}_{|T|}^{v}\right\}$ is given by 
\begin{equation}\label{8}
\hat{Y}^{v}=\boldsymbol{H}^{v} \boldsymbol{W}^{v}+ \boldsymbol{b}^{v}
\end{equation}
where $\boldsymbol{W}^{v} \in \mathbb{R}^{h2 \times 1}$, and $\boldsymbol{b}^{v}$ represents the bias.
}

Moreover, the bio-signal features are applied for the predictions of valence and arousal. In our model, we use the LSTM to model dependencies among these features. Let $\boldsymbol{X}^{e}=\left\{\boldsymbol{X}_{i}^{e} \in \mathbb{R}^{d_{e}}| i=1,...,|T|\right\}$ represents the ECG feature, $\boldsymbol{X}^{r}=\left\{\boldsymbol{X}_{i}^{r} \in \mathbb{R}^{d_{r}}| i=1,...,|T|\right\}$ represents the RESP feature and $\boldsymbol{X}^{b}=\left\{\boldsymbol{X}_{i}^{b} \in \mathbb{R}^{d_{b}}| i=1,...,|T|\right\}$ represents the BPM feature sequence. First, these features are concatenated by:
{
\begin{equation}\label{9}
\boldsymbol{X}^{B} = \operatorname{concat}\left(\boldsymbol{X}^{e}, \boldsymbol{X}^{r}, \boldsymbol{X}^{b}\right)
\end{equation}
where $d_{e}$, $d_{r}$ and $d_{b}$ represent the dimension of the ECG feature, RESP feature and BPM feature respectively. Afterwards, $\boldsymbol{X}^{B} \in \mathbb{R}^{T \times }$\\$^{(d_{e}+d_{r}+d_{b})}$ is sent to the LSTM and the output is given by:
\begin{equation}\label{10}
(\boldsymbol{H}_{1}^{B}, \ldots, \boldsymbol{H}_{|T|}^{B}) = \operatorname{L S T M}_{3}(\boldsymbol{X}_{1}^{B}, \ldots, \boldsymbol{X}_{|T|}^{B})
\end{equation}
where $\boldsymbol{H}_{i}^{B} \in \mathbb{R}^{h_{3}}$ and $h_{3}$ represents the dimension of $\operatorname{L S T M_{3}}$. Finally, the emotion prediction are made by a fully connected layer following the LSTM layer:
\begin{equation}\label{11}
\hat{Y}^{B}=\boldsymbol{H}^{B} \boldsymbol{W}^{B}+ \boldsymbol{b}^{B}
\end{equation}
where $\boldsymbol{W}^{B} \in \mathbb{R}^{h3 \times 1}$, and $\boldsymbol{b}^{B}$ represents the bias.
}

{\bfseries Late Fusion}: In this sub-challenge, we fuse multi-modal features using the late fusion strategy. First, $\hat{Y}^{a}$, $\hat{Y}^{v}$ and $\hat{Y}^{B}$ are concatenated by:
{
\begin{equation}\label{12}
\boldsymbol{Y}^{f} = \operatorname{concat}\left(\hat{Y}^{a}, \hat{Y}^{v}, \hat{Y}^{B}\right)
\end{equation}
where $\boldsymbol{Y}^{f}=\left\{\boldsymbol{Y}_{i}^{f} \in \mathbb{R}^{3}| i=1,...,|T|\right\}$. Susequently,  $\boldsymbol{Y}^{f}$ is sent to the LSTM model to get hidden states by:
\begin{equation}\label{eq13}
(\boldsymbol{H}_{1}^{f}, \ldots, \boldsymbol{H}_{|T|}^{f}) = \operatorname{L S T M}_{4}(\boldsymbol{Y}_{1}^{f}, \ldots, \boldsymbol{Y}_{|T|}^{f})
\end{equation}
where $\boldsymbol{H}_{i}^{f} \in \mathbb{R}^{h_{4}}$, and $h_{4}$ represents the dimension of $\operatorname{L S T M_{4}}$. Finally, the fusion emotion prediction $\hat{Y}^{f}=\left\{\hat{Y}_{1}^{f}, \ldots, \hat{Y}_{|T|}^{f}\right\}$ are made through a fully connected layer following the LSTM layer:
\begin{equation}\label{eq14}
\hat{Y}^{f}=\boldsymbol{H}^{f} \boldsymbol{W}^{f}+ \boldsymbol{b}^{f}
\end{equation}
where $\boldsymbol{W}^{f} \in \mathbb{R}^{h4 \times 1}$, and $\boldsymbol{b}^{f}$ represents the bias.
}

\subsection{MuSe-Physio sub-challenge}
In this subsection, we will describe the details of the method for the MuSe-Physio sub-challenge. Based on the model designed for the MuSe-Stress sub-challenge, the GCNN-LSTM-based model is added and used for modeling the temporal dependencies of acoustic feature and visual feature. It is noted that we use the VGGish feature instead of eGeMAPS feature in this sub-challenge. Figure 2 shows the architecture of our method, which contains four modules: the LSTM network with self-attention mechanism module for the acoustic and visual features, the gated CNN module with the LSTM network module for the acoustic and visual features, the LSTM module for the bio-signal features and the late fusion module. Firstly, we send the eGeMAPS feature and the VGGface feature to the LSTM network with self-attention module respectively and get the predictions from audio-visual modalities. Then, the same features are also sent to the gated CNN module with the LSTM network module respectively and get the predictions of audio and video modalities. Thirdly, the bio-signal features are concatenated and sent to the LSTM module in order to get the predictions of bio-signal modality. Finally, all of the predictions from above modalities are concatenated and sent to the late fusion module.\par

\begin{figure}[t]
  \centering
  \includegraphics[width=85mm]{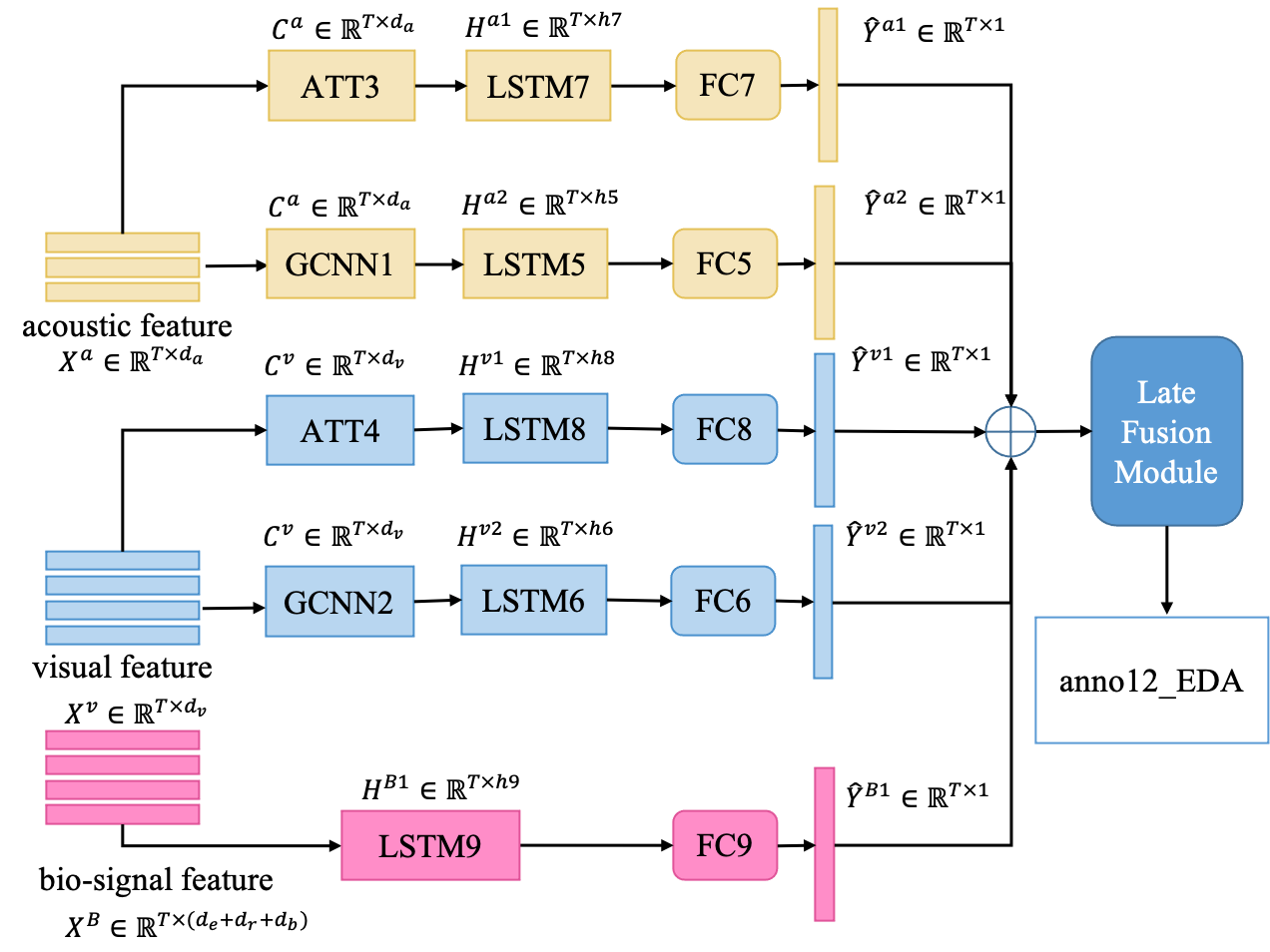}
  \caption{Overview of the architecture used in the MuSe-Physio sub-challenge. It contains four modules: the LSTM network with self-attention module for the acoustic and visual features, the gated CNN block with the LSTM network module for the acoustic and visual features, the LSTM module for the bio-signal features and the late fusion module.}
  \Description{}
\end{figure}

{\bfseries GCNN-LSTM-based Model}: In this sub-challenge, we first train the model consisted of stacked gated convolution blocks with a LSTM layer and a fully connected layer. For the gated block, the input acoustic sequence $\boldsymbol{X}^{a}$ to the output layer is defined as:
{
\begin{equation}\label{15}
\boldsymbol{G}^{a}=\operatorname{conv}(\boldsymbol{X}^{a}, \boldsymbol{W}^{a}) \odot \operatorname{sigm}(\operatorname{conv}(\boldsymbol{X}^{a}, \boldsymbol{Z}^{a}))
\end{equation}
where $\operatorname{sigm}$ is the sigmoid activation function, $\operatorname{conv}$ represents the convolution operation, $\odot$ is the Hadamard product between two tensors. $\boldsymbol{G}^{a} \in \mathbb{R}^{T \times d_{n}^{a}}$, $d_{n}^{a}$ is the number of channels for convolution. $\boldsymbol{W}^{a} \in \mathbb{R}^{d_{a} \times d_{n}}$ and $\boldsymbol{Z}^{a} \in \mathbb{R}^{d_{a} \times d_{n}}$ are parameters of the convolution layers. For simplicity, we override the whole operation as a mapping $\operatorname{GCNN_{1}}$:
\begin{equation}\label{16}
\boldsymbol{G}^{a}= \operatorname{GCNN_{1}}(\boldsymbol{X}^{a})
\end{equation}
}
Similarly, given the input visual sequence $\boldsymbol{X}^{v}$, the GCNN output is given by:
{
\begin{equation}\label{17}
\boldsymbol{G}^{v}= \operatorname{GCNN_{2}}(\boldsymbol{X}^{v})
\end{equation}
$\boldsymbol{G}^{v} \in \mathbb{R}^{T \times d_{n}^{v}}$, $d_{n}^{v}$ is the number of channels for convolution.
}
Then, the GCNN output is sent to the LSTM to capture the temporal dependencies within the feature sequences. Given the input embedding sequence $\boldsymbol{G}^{a}$, the LSTM output is given by:
{
\begin{equation}\label{18}
(\boldsymbol{H}_{1}^{a1}, \ldots, \boldsymbol{H}_{|T|}^{a1}) = \operatorname{L S T M}_{5}(\boldsymbol{G}_{1}^{a}, \ldots, \boldsymbol{G}_{|T|}^{a})
\end{equation}
where $\boldsymbol{H}_{i}^{a1} \in \mathbb{R}^{h_{5}}$ and $h_{5}$ represents the dimension of $\operatorname{L S T M_{5}}$. Similarly, given the input embedding sequence $\boldsymbol{G}^{v}$, we can get the LSTM output $\boldsymbol{H}^{v1} \in \mathbb{R}^{T \times h_{6}}$, where $h_{6}$ represents the dimension of $\operatorname{L S T M_{6}}$.
}\par
Finally, $\boldsymbol{H}^{a1}$ and $\boldsymbol{H}^{v1}$ are sent to a fully connected layer to make the emotion predictions respectively. According to the Eq. \ref{7}, $\hat{Y}^{a1}$ and $\hat{Y}^{v1}$ are calculated. $\hat{Y}^{a1}$ is the emotion predictions of audio modality and $\hat{Y}^{v1}$ is the emotion predictions of video modality.\par
In this sub-challenge, we also use the LSTM with self-attention mechanism to capture the complex temporal dependencies and make emotion predictions for audio and video modality. According to the Eq. \ref{1} -- \ref{8}, $\hat{Y}^{a2}$ and $\hat{Y}^{v2}$ are calculated. $\hat{Y}^{a2}$ is the emotion predictions of audio modality and $\hat{Y}^{v2}$ is the emotion predictions of video modality. In addition, the bio-signal features are also sent to the LSTM model and make emotion predictions. According to the Eq. \ref{9} -- \ref{11}, $\hat{Y}^{B1}$ is calculated. $\hat{Y}^{B1}$ is the emotion predictions of bio-signal modality.

{\bfseries Late fusion}: In this sub-challenge, we also fuse multi-modal features using late fusion strategy. $\hat{Y}^{a1}$, $\hat{Y}^{v1}$, $\hat{Y}^{a2}$, $\hat{Y}^{v2}$ and $\hat{Y}^{B1}$ are concatenated which is given by:
{
\begin{equation}\label{12}
\boldsymbol{Y}^{f1} = \operatorname{concat}\left(\hat{Y}^{a1}, \hat{Y}^{v1},\hat{Y}^{a2}, \hat{Y}^{v2}, \hat{Y}^{B1}\right)
\end{equation}
}
According to the Eq. \ref{eq13} -- \ref{eq14} we get the final emotion predictions $\hat{Y}^{f1}$.

\subsection{CCC Loss}
In both sub-challenges, we train our model using the Concordance Correlation Coefficient (CCC) loss, which can be formatted as follows:
{
\begin{equation}\label{12}
\mathcal{L}=1-C C C
\end{equation}
\begin{equation}\label{12}
C C C=\frac{2 \rho \sigma_{\hat{Y}} \sigma_{Y}}{\sigma_{\hat{Y}}^{2}+\sigma_{Y}^{2}+\left(\mu_{\hat{Y}}-\mu_{Y}\right)^{2}}
\end{equation}
where $\sigma_{\hat{Y}}$ and $\sigma_{Y}$ are the respective standard deviations. $\mu_{\hat{Y}}$ and $\mu_{Y}$ are the average of the predicted values $\hat{Y}$ and labels $Y$, correspondingly. $\rho$ is the Pearson Correlation Coefficient (PCC) between $\hat{Y}$ and $Y$.
}

\section{EXPERIMENTS}
\subsection{Dataset}
In MuSe 2021, two datasets (MuSe\_CaR \cite{dataset} and Ulm-TSST) are provided for four different sub-challenges. The dataset we used is Ulm-TSST, a novel subset of the audio-visual textual Ulm-Trier Social Stress dataset that features German speakers in a Trier Social Stress Test (TSST) induced stress situation.  This dataset contains 69 candidates (49 of whom are female), between the ages of 18 and 39, and providing a total of approximately 6 hours of data. The statistics information is shown in Table 1.

\subsection{Experimental Setup}
We implement all of our models with the PyTorch toolkit in the two sub-challenges. For the MuSe-Stress sub-challenge, the proposed model consists of the self-attention layer, a unidirectional LSTM layer and the fully connected layer. The number of heads is set to 4 or 8 and the number of layers is 1, 2, or 4. The number of hidden sizes in the unidirectional LSTM layer is 64, 128, or 256 and the number of the unidirectional LSTM layer is set 2 or 4. For the MuSe-Physio sub-challenge, the proposed model consists of a gated CNN layer, a unidirectional LSTM layer and the fully connected layer. The number of channels for convolution is 64 or 128 and the number of hidden size in the unidirectional LSTM layer is 128, 256, or 1024. The Adam optimizer with a learning rate (0.0002, 0.001, 0.002, or 0.005) is used to optimize the whole networks in both sub-challenges. During training, the batch size is set to (64, 128, or 256). We train the model 100 epochs. When the loss does not decrease in 15 consecutive epochs, the learning rate will be halved. For the late fusion model, we use a Bi-LSTM layer with 32 cells to fuse the previous predictions. The late fusion model is trained at most 20 epochs. The Adam optimizer with a learning rate of 0.001 is also applied, and the batch size is set to 64.

\begin{table}
	\caption{Statistics information of the Ulm-TSST dataset.}
	\label{tab:dataset}
	\begin{tabular}{c|cc}
		\toprule
		Partition&Number&Duration\\
		\midrule
		Train & 41 & 3 :25 :56\\
		Devel & 14 &1 :10 :50\\
		Test & 14 &1 :10 :41\\
		\midrule
		Sum & 69 & 5 :47 :27\\
		\bottomrule
	\end{tabular}
\end{table}

\subsection{Unimodal Results}
We first evaluate the performance of each modality we used in both sub-challenges. To verify the effectiveness of the proposed model, several experiments are conducted. The experiment results are given in Table \ref{tab:1} and Table \ref{tab:2}. From Table \ref{tab:1}, we can conclude that 1) the \lq{Self-Attn+LSTM}{\rq} model gets the best performances when predicting arousal and get a slight decrease when predicting valence. 2) When using the audio and video features to predict arousal and valence, the \lq{Self-Attn+LSTM}{\rq} model performs better than the \lq{LSTM}{\rq} model. It indicates that the proposed model can further model the dependency in the sequence. It can be seen from Table \ref{tab:2} that the \lq{GCNN+LSTM}{\rq} model can improve the CCC performance when using audio and video features. We can conclude that the \lq{GCNN+LSTM}{\rq} model is also suitable for dimensional emotion recognition. It also can be found that the \lq{Self-Attn+LSTM}{\rq} model achieves the best performance using the audio features. It shows the effectiveness of the self-attention mechanism. \par

\begin{table}
	\caption{ The CCC performance on the development set of the MuSe-Stress sub-challenge.
		\lq{M}{\rq} denotes the used modality.
	 \lq{A}{\rq}, \lq{V}{\rq} and \lq{B}{\rq} represent the audio, video, and bio-signal modality. }
	\label{tab:1}
	\begin{tabular}{cc|ccc}
		
		\toprule
		Feature&M&Model&Arousal&Valence\\
		\midrule
		eGeMAPS & A &Self-Attn&0.4158 &0.4123\\
		eGeMAPS & A &LSTM& 0.4304&0.5845\\
		eGeMAPS & A &Self-Attn+LSTM& 0.5142&0.5735\\
		\midrule
		VGGface & V &Self-Attn& 0.1412&0.4307\\
		VGGface& V &LSTM& 0.2004&0.4653\\
		VGGface & V &Self-Attn+LSTM& 0.2303&0.4543\\
		\midrule
		ECG+RESP+BPM & B&LSTM& 0.3921&0.2912\\
		\bottomrule
	\end{tabular}
\end{table}

\begin{table}
	\caption{The CCC performance of on the development set of the MuSe-Physio sub-challenge. 		\lq{M}{\rq} denotes the used modality. \lq{A}{\rq}, \lq{V}{\rq} and \lq{B}{\rq} represent the audio, video, and bio-signal modality.}
	\label{tab:2}
	\begin{tabular}{cc|cc}
		\toprule
		Feature&M&Model&anno12\_EDA\\
		\midrule
		VGGish & A &Self-Attn& 0.2391\\
		VGGish & A &LSTM& 0.3180\\
		VGGish & A &Self-Attn+LSTM& 0.5050\\
		\midrule
		VGGish & A &GCNN& 0.1486\\
		VGGish & A &GCNN+LSTM& 0.4400\\
		\midrule
		VGGface & V &Self-Attn& 0.3232\\
		VGGface& V &LSTM& 0.3903\\
		VGGface & V &Self-Attn+LSTM& 0.3745\\
		\midrule
		VGGface& V &GCNN& 0.1938\\
		VGGface & V &GCNN+LSTM& 0.4170\\
		\midrule
		ECG+RESP+BPM & B&LSTM& 0.4639\\
		\bottomrule
	\end{tabular}
\end{table}

\begin{table*}
	\caption{CCC performance of different modalities using late fusion strategy on the development set in the MuSe-Stress sub-challenge.}
	\label{tab:3}
	\begin{tabular}{c|c|cc}
		\toprule
		Feature&M&Arousal&Valence\\
		\midrule
		eGeMAPS + VGGface& A+V & 0.4922&0.6756\\
		eGeMAPS +ECG+RESP+BPM & A+B& 0.5378&0.5863\\
		VGGface + ECG+RESP+BPM & V+B & -0.0024&0.5649\\
		eGeMAPS + VGGface+ECG+RESP+BPM& A+V+B & 0.5422&0.6965\\
		\bottomrule
	\end{tabular}
\end{table*}

\begin{table*}
	\caption{CCC performance of different modalities using late fusion strategy on the development set in the MuSe-Physio sub-challenge.}
	\label{tab:4}
	\begin{tabular}{c|c|c|c}
		\toprule
		Feature&M&Model&anno12\_EDA\\
		\midrule
		VGGish + VGGface& A+V& Self-Attn+LSTM  & 0.5797\\
		VGGish + VGGface& A+V& GCNN+LSTM  & 0.5784\\
		VGGish + VGGface& A+V& Self-Attn+LSTM, GCNN+LSTM  & 0.6128\\
		\midrule
		VGGish + VGGface+ECG+RESP+BPM& A+V+B& Self-Attn+LSTM& 0.6153\\
		VGGish + VGGface+ECG+RESP+BPM& A+V+B& GCNN+LSTM& 0.5847\\
		VGGish + VGGface+ECG+RESP+BPM& A+V+B& Self-Attn+LSTM, GCNN+LSTM& 0.6257\\
		\bottomrule
	\end{tabular}
\end{table*}

\begin{table*}
	\caption{The best submission results of our proposed method in the two sub-challenges.}
	\label{tab:5}
	\begin{tabular}{c|c|c|c|c|c|c}
		\toprule
		Model&Sub-challenge&Partition&Arousal&Valence&Combined&anno12\_EDA\\
		\midrule
		Baseline&MuSe-Stress& Test& 0.4562 & 0.5614& 0.5088& -\\
		Proposed Model&MuSe-Stress& Test& 0.4609 & 0.6159& 0.5384& -\\
		\midrule
		Baseline&MuSe-Physio& Test& - & -& -& 0.4908\\
		Proposed Model&MuSe-Physio& Test& - & -& -& 0.5412\\
		
		\bottomrule
	\end{tabular}
\end{table*}

\subsection{Multimodal Results}
In our method, the late fusion strategy is used in both sub-challenge. Different from previous studies, the bio-signal features are also explored. Table \ref{tab:3} shows the CCC performance of different modalities using the late fusion strategy on the development set in the MuSe-Stress sub-challenge. From this table, we can find that 1) the multi-modal features can get better performance than the uni-modal features. 2) the bio-signal features fusing with the audio and video features prove to be effective, it can further improve the model performance based on the fusion of audio and video features. 3) the bio-signal features seem not to improve the model performance when only fusing with the video features. The best results for predicting arousal and valence are 0.5422 and 0.6965 respectively. Table \ref{tab:4} shows the CCC performance of different modalities using the late fusion strategy on the development set in the MuSe-Physio sub-challenge. From this table, we can see that 1) the bio-signal features fusing with other features can further improve the model performance. 2) different models can also improve the CCC performance when using the same features. The best result for predicting anno12\_EDA is 0.6257, which outperforms the baseline (0.4913) and the best uni-modal result (0.5050).
\subsection{Submission Results}
Table \ref{tab:5} shows the best submission results of the proposed method in the two sub-challenges. For the MuSe-Stress sub-challenge, our proposed method outperforms the official baseline by 0.0047 with arousal and 0.0545 with valence. It is noticed that this sub-challenge is an extremely challenging task. There is a heavy overfitting problem here, especially predicting arousal. For the MuSe-Physio sub-challenge, the proposed method outperforms the official baseline by 0.0504. Before submitting the best results, we retrain the data from the development set using the optimal parameters in this sub-challenge, in order to further enhance the generalization ability of the model.
\section{CONCLUSIONS}
In this paper, we present our solutions for the MuSe-Stress sub-challenge and the MuSe-Physio sub-challenge of Multi-modal Sentiment Challenge (MuSe) 2021. For the MuSe-Stress sub-challenge, fusing the bio-signal features with the audio features(eGeMAPs) and the visual features (VGGface) proved useful for predicting arousal and valence. Also, the LSTM network as well as the self-attention mechanism and late fusion strategy can further improve the model performance. For the MuSe-Physio sub-challenge, the bio-signal features are explored and proved to be useful. Besides, the LSTM network as well as the self-attention mechanism and the GCNN-LSTM based model are used for this task and achieve good performance. The late fusion strategy is also applied and contributes mostly to the model performance.\par
The proposed method shows promising perspectives of future improvements. First, more multi-modal features can be used in both sub-challenge. Then, many advanced and robust models can be explored in the next step. Finally, the early fusion strategy is proved useful, and we can use it to the two sub-challenges.

\section{Acknowledgments}
This work is supported by the National Key Research and Development
Project of China (2018YFB1305200) and the National Natural Science Fund of China (62171183, 61801178).



\bibliographystyle{ACM-Reference-Format}
\balance
\bibliography{sample-base}

\appendix

\end{document}